\documentclass[conference]{IEEEtran}

\usepackage[utf8]{inputenc} 
\usepackage[T1]{fontenc}    
\usepackage{hyperref}       
\usepackage{url}            
\usepackage{booktabs}       
\usepackage{amsfonts}       
\usepackage{color} 
\usepackage[super]{nth}
\usepackage{calc}
\usepackage{float}
\usepackage{multirow}
\usepackage{amsmath}
\usepackage{algorithm2e}
\usepackage{algorithmic}
\usepackage{mathtools} 
\usepackage{bm} 

\usepackage{multicol}
\usepackage{subfig}

\usepackage{tikz}
\definecolor{mycolor1}{rgb}{1,0.2,0.3}
\definecolor{mycolor2}{rgb}{0.2,0.3,1}
\tikzstyle{tau1} = [mycolor1, dashed]
\tikzstyle{tau2} = [mycolor2, densely dashdotted]
\tikzset{cross/.style={path picture={ 
  \draw[black]
(path picture bounding box.south east) -- (path picture bounding box.north west) (path picture bounding box.south west) -- (path picture bounding box.north east);
}}}
\definecolor{Gray}{gray}{0.9}
\definecolor{LightCyan}{rgb}{0.88,1,1}
\definecolor{Orange}{rgb}{1,0.5,0.2}
\definecolor{Yellow}{rgb}{1,1,0}
\definecolor{Green}{rgb}{0.5,0.9,0}
\definecolor{Blue}{rgb}{0,0.7,1}
\definecolor{LtBlue}{rgb}{0.4,0.8,1}
\definecolor{DkBlue}{rgb}{0., 0.447,0.69}
\definecolor{Purple}{rgb}{.7,0.5,1}

\newcommand{\eqdef}{\overset{\text{def}}{=}}

\DeclarePairedDelimiter\abs{\lvert}{\rvert}%
\DeclarePairedDelimiter\norm{\lVert}{\rVert}%
\makeatletter
\let\oldabs\abs
\def\abs{\@ifstar{\oldabs}{\oldabs*}}
\let\oldnorm\norm
\def\norm{\@ifstar{\oldnorm}{\oldnorm*}}
\makeatother

\usepackage{todonotes}

\title{Optimization of computational budget for power system risk assessment}

\author{\textbf{Benjamin Donnot$^\ddagger$ $^\dagger$}\thanks{Benjamin Donnot corresponding authors: benjamin.donnot@inria.fr}, \textbf{Isabelle Guyon$^\ddagger\bullet$},  \textbf{Antoine Marot$^\dagger$ }, \textbf{Marc Schoenauer$^\ddagger$}, \\, \textbf{Patrick Panciatici$^\dagger$} \\
  $\ddagger$ UPSud and Inria TAU, Université Paris-Saclay, France. \\
 $\bullet$ ChaLearn, Berkeley, California.  $\dagger$ RTE France.}

\definecolor{mycolor1}{rgb}{1,0.2,0.3}
\definecolor{mycolor2}{rgb}{0.2,0.3,1}
\definecolor{mycolor3}{rgb}{0.,0.4,0.}

\newcommand{\distas}[1]{\mathbin{\overset{#1}{\kern\z@\sim}}}%
\newsavebox{\mybox}\newsavebox{\mysim}
\newcommand{\distras}[1]{%
  \savebox{\mybox}{\hbox{\kern3pt$\scriptstyle#1$\kern3pt}}%
  \savebox{\mysim}{\hbox{$\sim$}}%
  \mathbin{\overset{#1}{\kern\z@\resizebox{\wd\mybox}{\ht\mysim}{$\sim$}}}%
}

\begin{document}

\maketitle

\begin{abstract}
We address the problem of maintaining high voltage power transmission networks in security at all time, 
namely anticipating exceeding of thermal limit for eventual single line disconnection (whatever its cause may be) by running slow, but accurate, physical grid simulators. 

New conceptual frameworks are calling for a probabilistic risk-based security criterion. However, these approaches suffer from high requirements in terms of tractability. Here, we propose a new method to assess the risk. This method uses both machine learning techniques (artificial neural networks) and more standard simulators based on physical laws. More specifically we train neural networks to estimate the overall dangerousness of a grid state. A classical benchmark problem (manpower 118 buses test case) is used to show the strengths of the proposed method.
\end{abstract}

\section{Problem definition}

Today's European power-grids are facing new challenges. Renewable energies such as wind and solar power play an increasing role in the production. The electricity market is growing, and the total demand has stopped increasing. All these factors combined make the task more difficult for TSOs.
In the past, the rise in complexity was managed by building new heavy infrastructures required by the growth of consumption. This is not possible anymore since growth in revenue is stalling as well. TSO's need to be addressed by optimizing current infrastructures, finding new flexibilities. 

It is then becoming more critical for TSOs today to move away from reactive real time grid management to an approach based on anticipation with real time automation.
This means that TSO's should 
make studies in anticipation, with grids states coming from forecast available at the date and time of the study. This introduce even more variability in the grids states. One way to tackle of these uncertainties is to use a Monte Carlo approaches presented in the ITESLA\footnote {See \href{http://www.itesla-project.eu/itesla-results.html}{http://www.itesla-project.eu/itesla-results.html} for more information.} framework. This means simulating a lot of possible grid states, and increasing the computation needed to assess the security. This computational need is also induced by the GARPUR methodology, where a stochastic security criterion has been defined.

In this paper, we will (1) study the risk of a given grid state (see section \ref{sec:math} equation \ref{eq:risk} for a formal definition of this risk), (2) propose a method to rank contingencies in decreasing severity, (3) evaluate the potential cost of \emph{not} simulating a set of contingencies $\mathcal{V}$ (what is called "residual risk" in the next sections) and (4) to propose a way to mix regular approaches and machine learning to increase computational speed without sacrificing accuracy


Other authors use machine learning to address power system related problems. In these papers, most of the time people try to classify grid state according to some security criteria (\cite{wehenkel1997machine}, \cite{wehenkel2012automatic}, \cite{saeh2008static}, \cite{fliscounakis2013contingency}), or to predict how a system will react after an unplanned event occur (\cite{Duchesne2017}). We believe our approach to be different: we learn how to rank contingencies in order to run physical simulators on a limited accurate amount of situations

Our proposed method relies on previously published work in \cite{bonnot:hal-01581719} and \cite{donnot:hal-01695793} in which we devised a neural-network trained with "guided dropout", to predict power flows in power grids for given topology variants while training only on a small subset of these. We pursue the evaluation of this strategy in this paper, where neural networks are solely trained on a small set of configurations (less than $1\%$).

\section{Statement of the problem and notations \label{sec:math}}

Under our assumptions, a ``system state'' $x$ consists of the power flowing in all lines, resulting from given (fixed) injections, for a specific grid topology. We always analyze a situation corresponding to a fixed state in this paper and sometimes omit $x$ for brevity of notation. We also omit to specify time ordering, although states are time ordered. 
A contingency $z$ might arise with probability $p(z)$
and is associated with a loss function $L(z;x)$. For instance, events $z$ might be single line disconnections occurring with probability $p(z)=\pi(1)$ or double line disconnections occurring with probability $p(z)=\pi(2)=\pi(1)^2$ (thus assuming that two disconnections are independent). The overall risk is defined as:
\begin{equation}
R_{\max}(x) = \sum_{z \in \mathcal{Z}} p(z) L(z;x) \label{eq:risk}
\end{equation}
\noindent This definition is like the one presented in \cite{karangelos2016probabilistic} Eq. 3, for instance. In our application context, we assume that $L(six)$ is the $\{0, 1\}$ loss, with $0$ meaning that the contingency $z$ arising in state $x$ is innocuous and $1$ that it is risky or "dangerous" for our system ({\em i.e.} at least one line, still in service after $z$ arose, will exceed its thermal limit). Thus:

\begin{equation}
  L(z;x) =  \left\{
    \begin{aligned}
      0 ~&~\text{``No current flowing on any line} \\
         &~\text{exceeds the line thermal limit } \\
         &~\text{under contingency $z$} \\
         &~\text{in grid state $x$''} \Rightarrow \text{OK}\\
      1 ~&~ \text{``Otherwise''} \Rightarrow \text{``Bad'' contingencies}
    \end{aligned}
  \right.
\end{equation}
\noindent Estimating the real damage of the grid would endure after contingency $z$ would require computing the real  "real" behaviour of the grid including corrective actions, load shedding and a full "cascading failure" (as presented in \cite{1508.01775} for example), which is computationally too expensive to calculate presently.

We could also refine this loss $L(z;x)$ in multiple fashion. We could for instance take into account the depth of congestions. One step further, we could use ITESLA Methodology to further take into account the flexibilities on the grid and classify contingencies in 4 categories: 1) not dangerous 2) dangerous, but corrective action can be implemented to restore the security 3) dangerous, but there exists a preventive action that can still be taken to cure the grid 4) dangerous, and no known solution to cure the grid exists. 

Our approach is different: our loss can be interpreted as "$L(z;x)=0$: no need to manually study the contingency, the grid is safe" and "$L(z;x)=1$: the contingency is not safe, a more precise evaluation of its impact must be carried out, either manually, or with the more accurate simulator". We believe this approach of mixing machine learning for ranking, 
and physical simulation for grid state evaluation is promising and show it in the section \ref{sec:results}.

As we already explained, the computational budget needed to be performed will drastically increase in the near future
Because of computational costs, we cannot carry out all the computation needed as explained in the ITESLA project see \footnote{See the ITELSA project at \href{http://www.itesla-project.eu/index.html}{http://www.itesla-project.eu/index.html}.}). Hence, we will use a "fast proxy"\footnote {Using neural network with dedicated architecture will shows that a speed-up of more than $1.000$ is achievable compare to actual load-flow simulators.} that will rank the contingencies $z \in \mathcal{Z}$ to focus our computational budget (number of call to the physical simulator) on the most critical one. 


If we evaluate with the physical simulator a set $\mathcal{V}$ of contingencies, the \emph{residual} risk corresponding to events in $\mathcal{Z} - \mathcal{V}$ is:
\begin{equation}
R(\mathcal{V}; x) = \sum_{z \in \mathcal{Z} \-- \mathcal{V}}   p(z) L(z;x) \label{eq:res_risk}
\end{equation}
This corresponds to the risk taken of \emph{not} computing (with the physical simulator) the contingencies in $\mathcal{Z} \-- \mathcal{V}$ with the slow simulator. This residual risk $R$ is bounded between:
\begin{equation}
	R(\mathcal{Z}; x)=0 \text{~~~and~~~}   R(\emptyset; x)=R_{\max}(x)
\end{equation}

In this paper, because we use a benchmark of modest size, we can exhaustively compute $L(z;x)$ for all $z$ with the physical simulator to presents results. In practice $R(\mathcal{V}; x)$ might have to be approximated by replacing $L(z;x)$ with an approximate loss $\hat{L}(z;x)$, obtained using power flows estimated by our ``proxy'' simulator (precisely defined in section \ref{sec:method} equation \ref{eq:lhat}).

\section{The power grid problem}
In this paper, we consider only two kinds of contingencies: ``single contingency'', denoted by $z_i$, representing the disconnection of one single power line, and ``double contingency'' $z_{i,j}$ representing the disconnection of two lines.

After the power grid suffered a single contingency, we will say its state is in "n-1" 
If $n$ denotes the number of lines in our power grid, there are exactly $n$ different "n-1" grid states. Similarly, a power grid suffering a double contingency will be referred to as a "n-2". 

\subsection{Beyond "N-1" security policy}

Commonly, TSO operate the  grid using the so called "N-1" security policy. This policy stipulates that should \emph{ANY} unplanned single contingency occurs, the flow on all the lines of the power grid must remain below their thermal limits, or be set back by a curative action within an authorized short time window 
. This terminology should not be confused with the "n-1" state in which the grid finds itself after one line disconnection. In fact assessing the "N-1" security requires computing at least $n$ load-flows, each one corresponding to one possible "n-1" grid state. 

For example, the French power grid counts approximately $n \approx 10 000$ power lines. Thus, assessing the "N-1" security of this network requires  $\approx 10 000$ load-flows, and assessing the "N-2" security would require on the order of $\frac{n(n-1)}{2} \approx 50~10^6$ load flow evaluations. In this context, it is understandable given a computation budget near real time
that TSO's do not operate under higher order security policies, such as "N-2" (two line disconnections), "N-3", and so on, which most of the time have very low probability. 

One of our motivation for studying "N-2" grid safety is that TSO operators must anticipate future grid states on an ever longer horizon to guarantee security as we have already developed in previous sections so that there is time to test if a remedial actions can be taken in real time.  Our method would allow to gain speed in evaluating grid security. This could allow TSO to reduce cost by a better anticipation of the risk, or increase the security, with a given budget.

During the training the neural network sees only states where at most one power line is disconnected (single contingencies). As the neural network never sees at training time, states where $2$ power lines where missing, evaluating "N-2" security
is an effective way to evaluate how well our estimation will perform in unseen scenarios (when grid states it is tested on differs from what it learns). This is really important in practice. For such critical systems as the power grid, we must make sure the method does not lead to taking bad decisions when facing unseen configurations.

Given their really low probability of occurrence, we ignore the effects (and the residual risk) associated with higher order contingencies ("n-3", "n-4", etc.). As a further simplification, we assume that all single disconnections $z_i$ have equal probability:
  	\begin{equation} p(z_i) \eqdef \pi_{(1)}\end{equation}
and all double disconnection $z_{i,j}$ have equal probability:
  	\begin{equation} p(z_{i,j}) \eqdef \pi_{(2)}  = \pi_{(1)}^2. \label{eq:proba_emp}
    \end{equation}
In reality, such probabilities vary depending on factors such as line length, pair of line proximity, local climate, weather variations, etc. Such variations are neglected in the present paper, but can make area of future studies.

\subsection{Parameters setting}
In this section we expose how we choose the parameters of our experiment to be as realistic as possible of the French power grid. 

Expert dispatchers (TSO operators responsible of the grid security) estimate that a full "N-1" simulation yields approximately $100$ "bad" events (dangerous contingencies) for the peak total demand. This means that approximately $1\%$ of the single events should present a serious risk requiring a corrective action. To respect the order of magnitude of this proportion of "bad" events, we used a calibration dataset, which allowed us to \emph{set} the thermal limits
of each line in our test case grid. Having set these values, we evaluate them on the full "N-1" for $100$ different grid states\footnote{See the section \ref{sec:data} for a detailed description of this dataset.}, requiring to compute $18~600$ load flows. Among all the "n-1" events investigated, $1.75\%$ were found unsafe in our simulations and $4.22\%$ for "n-2" events. 

We want our study to be representative of the behavior of the French power grid. we also used real data coming from the French power grid starting January \nth{1} 1994 to December \nth{31} 2015 to have estimates of $\pi_{(1)}$ and  $\pi_{(2)}$ on the French power grid. We avoided choosing time segments during which catastrophic events occurred\footnote{For example the "Lothar" windstorm of December \nth{26} 1999, "Martin" windstorm of December \nth{27}-\nth{28} 1999 or the "Klaus" windstorm of January \nth{23}-\nth{24} 2009}. Indeed they are not particularly relevant for our study and would lead to overestimating these probabilities. Our dataset contains the dates and times of all failures of RTE material during the time segment chosen. We estimate the probability of single failure per hour as:
\begin{equation}
\pi_{(1)}^{\text{FR}} = \frac{n_f}{n_h.n}
\end{equation}
\noindent where $n_f$  and $n_h$ are the number of failures and the number of hours in our dataset respectively, and $n$ the number of power lines.
This gives us an estimate of the French single line failure probability:
\begin{equation}
	\pi_{(1)}^{\text{FR}} \approx 1.2~10^{-6}
\end{equation}
\noindent This means that, on average, a given powerline $i$ will fail nearly once every $10^{6}$ hours\footnote{This is more than $100$ years.}. With the same technique, we found that:
\begin{equation}
	\pi_{(2)}^{\text{FR}} \approx 2.2~10^{-11}
\end{equation}

In this paper, we consider a smaller test case counting only $186$ power lines (instead of $n^{\text{FR}}\simeq 10~000$ for the French powergrid). Using the same probabilities for this smaller test case would lead to greatly underestimate the residual risk associated with the double contingencies. 

This led us to make adjustments to these probabilities. Let's consider the worst possible case, where all the contingencies are bad, to have an upper bound on the risk. In the French power system, the residual risk associated with all the "N-1" contingencies is $\pi_{(1)}^{\text{FR}}  \times n^{\text{FR}} \approx 0.12 $
and the risk associated with all the double  contingencies is $\pi_{(2)}^{\text{FR}}  \times \frac{n.(n-1)}{2} \approx 0.06$. Keeping the ratio "risk N-1 / risk N-2" constant across the grid state yields to consider:
\begin{equation}
 \frac{\pi_{(1)} \times n}{\pi_{(2)} \times \frac{n (n-1)}{2}} = \frac{\pi_{(1)}^{\text{FR}} \times n^{\text{FR}}}{\pi_{(2)}^{\text{FR}} \times \frac{n^{\text{FR}} (n^{\text{FR}}-1)}{2}}  \approx \frac{0.12}{0.06}
\end{equation}
\noindent Together with the assumption made in equation \ref{eq:proba_emp} and using $n=186$ (the size of our test case grid), we obtain :
\begin{align}
    \pi_{(1)} & \approx 5.4 ~ 10^{-3} \\
    \pi_{(2)} & \approx 2.9 ~ 10^{-5}
\end{align}

If these scaling where not performed, the residual risk associated with all the double contingencies in our test cases could be completely neglected compare to the risk associated with single ones. There would not be any advantages of using machine learning, as the accumulated risk of all the double contingencies would be almost zero\footnote{It would be of $\approx 10^{-7}$ if \emph{ALL} the double contingencies causes security problem, compare to $\approx 10^{-6}$ if \emph{ONLY ONE} single contingencies causes problem.}. This would lead to underestimate this risk associated to the double contingencies on the French power grid artificially (due to the size of our test case).

\section{Proposed methodology}
\label{sec:method}
Suppose we have a high end slow simulator and low end fast proxy simulator. We would like to take the most of them by combining them in a smart way, to best estimate the risk given an available computational budget. To do so, we first compute an estimate of the loss of contingency $z$ on grid state $x$ (denoted above $L(z;x)$). Then we are able to have an unbiased estimator of its severity score $p(z).L(z;x)$ (eg the loss scaled with the probability of the contingency).


Considering a fixed grid state $x$ and a given contingency $z$, we denote by $f_i$ the flow, computed with the high-end simulator, on the $i^{\text{nth}}$ line of grid $x$ after contingency $z$ occurs, and by $\bar{f}_i$ the thermal limit for this line.

We propose to first train a neural network with ``guided dropout'', as describe in \cite{donnot:hal-01695793} to approximate rapidly the power flow for the given grid state $x$ and contingency $z$. During the training step, only single contingencies are seen by the neural network.

Once the neural network is trained, we use it to predict flows. $\hat{f}_i$ denotes the flow predicted by our proxy (in this case our neural network) for the $i^{\text{nth}}$ line of the power grid.

It has been observed that neural networks tend to be "over confident" in their predictions (see for example \cite{DBLP:journals/corr/NguyenYC14}). This overconfidence could lead to a bad ranking with dramatic effects in practice. We propose to calibrate the score of our neural network by taking into account a fixed (yet calibrated) uncertainty by assuming:
\begin{equation}
	\forall i, (f_i-\hat{f}_i)  \sim \mathcal{N}\left(0, \sigma_i\right)
	\label{eq:error_normality}
\end{equation}
\noindent where $\sigma_i$  represents the model uncertainty for line $i$. This is a really simplistic model of the error of our model but this assumption is often made in practice, and sufficient for our needs here, as shown in section \ref{sec:results}. In the presented experiments, we calibrate the vector $\bm{\sigma}$ (of dimension $n$) using a calibration set distinct from the training set.
For real time operation, this vector $\bm{\sigma}$ can be calibrated using grid states available in real time, but for which the neural network has still not be trained on\footnote{the operator would still perform the full "N-1" computation, and this computations can be used to calibrate this vector}.

On this calibration set, we compute the true values $f_i$, using the high-end simulator, and the predictions $\hat{f}_i$ coming from our proxy: $\sigma_i$ is set to:
\begin{equation}
  \sigma_i \eqdef \frac{1}{\text{number of simulations}} . \sum_{\text{simulation} s} (\hat{f}_i - f_i)^2
\end{equation}

These $\sigma_i$'s are then used to compute the scores $\hat{L}_i$ that a given line is above its thermal limit as:
\begin{equation}
  \hat{L}^{(\text{aux})}_i \eqdef 1-F_{\sigma_i}(\bar{f}_i-\hat{f}_i)
\end{equation}
\noindent where $F_{\sigma_i}$ is the cumulative density function of the Normal law with mean $0$ and variance $\sigma_i$. 
This is equivalent to computing the "p-value" in the statistical test "$\hat{f}_i \leq \bar{f}_i$ supposing the error are normally distributed (eg supposing equation \ref{eq:error_normality}). For our problem, a grid is said to be ``non secure'' after contingency $z$, if at least one of its line is above its thermal limit. The score of the power grid, in state $x$ after contingency $z$, is then obtain with:
\begin{equation}
	\hat{L}^{(\text{aux})}(z; x) ~\eqdef~ \max_{1\leq i \leq n}  \hat{L}^{(\text{aux})}_i(z;x)
\end{equation}

This estimator $\hat{L}^{(\text{aux})}(z; x)$ is a biased stochastic estimator of the true risk $L(z; x)$: $\mathbb{E}\left( \hat{L}^{(\text{aux})}(z; x) \right) \neq L(z; x)$. We use the same calibration set to evaluate:
\begin{equation}
b \eqdef \mathbb{E}_{\text{calibration set}}\left( \hat{L}^{(\text{aux})}(z; x) \right) - L(z; x)
\end{equation}
\noindent We then finally obtain the unbiased estimator of the severity of the contingency $z$ on situation $x$:
\begin{equation}
\hat{L}(z; x) \eqdef \max \{ \hat{L}^{(\text{aux})}(z; x) - b, 0\} \label{eq:lhat}
\end{equation}
\noindent This "evaluated loss" $\hat{L}(z; x)$ is an unbiased estimator of the loss of the contingency $z$: $L(z; x)$. An estimator of the severity score of contingency $z$ is then
\begin{equation}
	\hat{s}(z;x) \eqdef \hat{L}(z;x).p(z)
\end{equation}
This severity score $\hat{s}(z;x)$ is an estimator of the impact of not computing the contingencies $z$ on the total risk defined in equation \ref{eq:risk}: it is the estimate of $L(z;x).p(z)$. Contingencies are ranked according to their respective severity score $\hat{s}(z;x)$: we want first to simulate contingencies that can cause the highest damage (pondered with the probability of occurrence). And the associated empirical maximum risk $\hat{R}_{\max}(x)$, and empirical residual risk $\hat{R}(\mathcal{V}, x)$ are defined with
\begin{align}
  \hat{R}_{\max}(x) & \eqdef \sum_{z \in \mathcal{Z}} p(z) \hat{L}(z;x) \label{eq:est_risk}\\
  \hat{R}(\mathcal{V}; x) &\eqdef \sum_{z \in \mathcal{Z} \-- \mathcal{V}}   p(z) \hat{L}(z;x) \label{eq:res_risk_est}
\end{align}
In analogy of section \ref{sec:method}, $\hat{R}_{\max}(x)$ is an estimatation of $R_{\max}(x)$, the overall risk of the situation\footnote{eg this is an estimate of the risk define in the GARPUR framework, see \cite{karangelos2016probabilistic}.} and $\hat{R}(\mathcal{V}; x)$ is an approximation of $R(\mathcal{V}; x)$. It represents the risk of \emph{not} computing with the physical simulator the contingencies \emph{not} in $\mathcal{V}$.

\section{Results \label{sec:results}}
In this section, we will show the results of the experiments carried out: we conduct systematic experiments on small size benchmark grids from Matpower \cite{Zimmerman11matpowersteadystate}, a library commonly used to test power system algorithms \cite{alsac1974optimal}. We report results on the largest case studied: a 118-nodes grid with $n=186$ lines.

We generate $500$ different grid states changing the injections $x$ of the initial grid given in Matpower. To generate semi-realistic data,  we used our knowledge of the French grid, to mimic the spatio-temporal behavior of real data~\cite{bonnot:hal-01581719}. For example, we enforced spatial correlations of productions and consumptions and mimicked production fluctuations, which are sometimes disconnected for maintenance or economical reasons. Target values were then obtained by computing resulting flows in all lines with the \emph{AC} power flow simulator \emph{Hades2}. 
 
On these $500$ cases, we then computed, still using the high-end simulator Hades2, the full "N-1" (making $500 \times 186 = 93~000$ load flow computations). Among this dataset, $75\%$ have been used for training our model, and the rest ($25\%$) for finding the best architecture and meta-parameters (learning rate, number of units per layer, number of layers, etc.) for the neural networks.
We note that, to be able to estimate the overall generalization of our method, we don't train our neural network on double contingencies.\\

For the calibration dataset, we simulate $100$ different grid states $x$, and the full "N-1" and "N-2" for all of these simulations. The test set is also composed of $100$ different grid states, and their full "N-1" and "N-2". The grid states in the test set are different from the one of the calibration set and the one in the training / validation set, and have never been seen during either training, or parameters estimation. We also want to emphasize that the distributions of the test set (representing the data the network will be tested on) and the distribution of the training set (data available for training the model, corresponding to what operators do today) are different: the test set is composed of single and double contingencies whereas the training counts only single contingencies.

The first section presents results in the estimation of the total risk $R_{\max}$ relying solely on machine learning, with almost no computing cost. In the second subsection, we show how the estimation of the most dangerous contingencies using the slow simulator can improve these approximations.

\begin{figure*}[ht!]
\centering
\subfloat[][]{\includegraphics[width=0.43\linewidth]{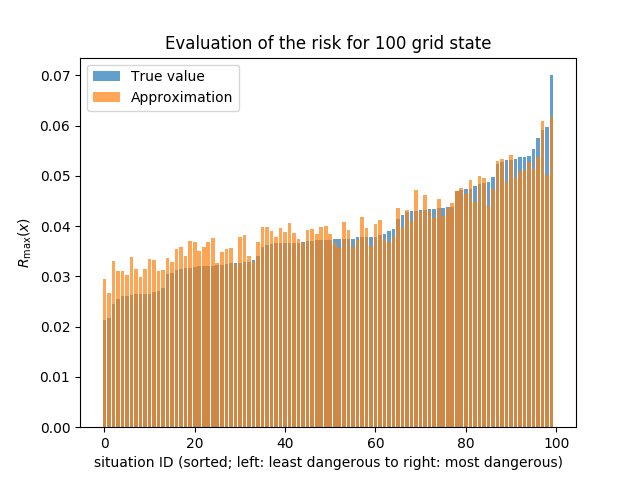} \label{fig:hist_risk}}
\subfloat[][]{\includegraphics[width=0.43\linewidth]{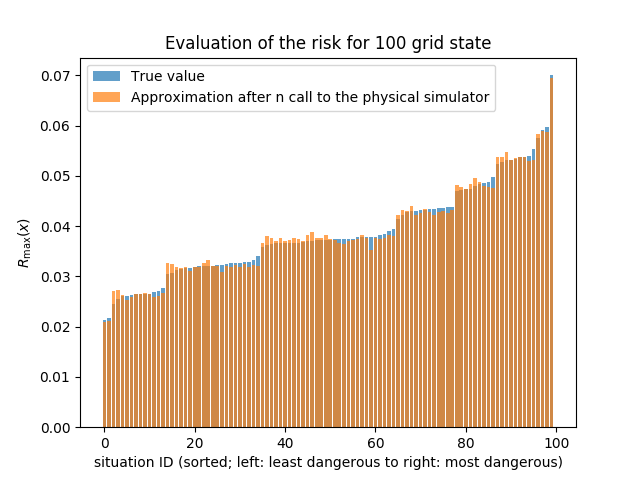} \label{fig:hist_risk_withcall}}
\caption{{\bf Histogram representing the total risk}, in is orange the empirical risk $\hat{R}_{\max}$ defined in section \ref{sec:method}.  In blue is the true total risk $R_{\max}$ for the $100$ situations of the test set (a) when relying solely on machine learning (see section \ref{sec:ml_only}) and (b) on allowing $n=186$ calls to the physical simulator (see section \ref{sec:ml_ps}).}
\end{figure*}

\subsection{Estimation of the total risk $R_{\max}$ relying on machine learning only}
\label{sec:ml_only}

The figure \ref{fig:hist_risk} presents the risk of the $100$ situations of the test set: the true risk $R_{\max}$ is represented in blue and is computed with the physical simulator according to the equation \ref{eq:risk}\footnote{This is not available in practice as it would require too much call to the physical simulator.}, and in orange the estimated risk $\hat{R}_{\max}$, defined in equation \ref{eq:est_risk}\footnote{This is the evaluation of the risk using the fast proxy alone, that come almost for free - a speed up of more than a 1000 is achievable in first experiments.}. In operational processes, the true risk $R_{\max}$ is unknown. For clarity in the representation, the $100$ test set situations have been sorted  in increasing order of $R_{\max}$. 

As we can see on the figure \ref{fig:hist_risk} (left), an estimate of the overall risk is possible. Our estimate $\hat{R}_{\max}$ is quite close on average of the total risk $R_{\max}$. The MAPE\footnote{Mean Absolute Percentage Error, define for two vector $x$ and $y$ of size $m$, $MAPE(x,y) = \frac{1}{m}.\sum \abs{\frac{x_i-y_i}{y_i}}$.} is $8.7\%$: Globally, we are also able to predict which situations will be the riskiest: the Pearson correlation coefficient between the estimate and the true values is $0.96$: there exists almost a linear relation between the proposed estimate and the actual true value.

But, for the most interesting cases, where the true risk is the highest, the performance decreases. In the most interesting cases for the TSO, the empirical risk estimation is bellow the true risk, which can be misleading, and is not suitable.

This estimation of the risk only relies on machine learning. This has limitation as we just exposed. In the next subsection, we will expose how a careful use of a physical simulator can increase the precision of the estimation of the risk.

\subsection{Estimation of the residual risk $R(\mathcal{V})$ with machine learning}
\label{sec:ml_ps}
In this section we will propose a second method, that will combine machine learning and physical simulators to estimate the overall dangerousness of a grid state.

In practice, the proposed methodology allows to rank the contingencies in decreasing order of risk (according to the method describe in section \ref{sec:method}). In real time, we can rely on the slow simulator to study carefully the riskiest ones. And this is the whole idea behind the "residual risk": riskiest situations are studied with physical simulators and the others are not.

Let's first consider the top $n$ (recall that $n$ is the number of power lines) contingencies that are simulated with the physical simulator. This is representative of what operators do today when they compute the full "N-1".  On the contrary of what operator do today 1) our strategy does not rely on simulating always the same kind of contingencies (all the single contingencies in today's operational processes) 2) use machine learning to evaluate the residual risk $R(\mathcal{V})$. 

In this framework, the overall risk can be evaluated as being: the true risk $p(z).L(z;x)$ for the "top n" contingencies ranked according to the results of the neural network (see section \ref{sec:method}). We then add the empirical residual risk $\pi(z).\hat{L}(z;x)$ for all the other contingencies.  The results for this new estimate of the risk are presented in figure \ref{fig:hist_risk_withcall} (right). As we can see, there is a significant improvement. Using a slow simulator can drastically help increasing the precision of the risk. The MAPE between this new estimates and the real value is $2.5\%$ compares to $8.7\%$ with the machine learning only. 

\begin{figure}[ht!]
\centering
  \includegraphics[width=0.8\linewidth]{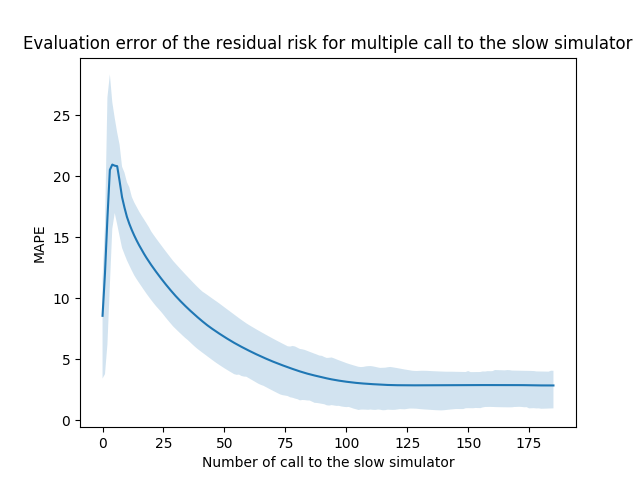}
  \caption{{\bf Representation of the error (MAPE) in the residual risk estimation} for the $100$ grid states of the test set as function of the number of calls to the physical simulator. The error bar represents the [25\%-75\%] confidence interval.}
  \label{fig:mapeforslowsimul}
\end{figure}

This phenomenon can be explained. The estimation of the residual risk is easier than the one of the total risk as shown in the figure \ref{fig:mapeforslowsimul}. This figure presents the error between the estimated residual risk (available in real time) and the true residual risk, as function of the number of calls to the physical simulator. For measuring the error, we choose to use the MAPE (defined in the previous subsection). We zoomed the plot on the interesting part for the TSO when the number of cal to the slow simulator is less than $n$ (the number of line in the power grid). This corresponds to actual operational processes, when operators compute the full "N-1".

As we can see on the figure \ref{fig:mapeforslowsimul}, the error on the residual risk decreases after a few calls to the simulator. This is not proper to the error used, the same shape is obtained when considering other error measures (such as the Root Mean Squared Error). The error is divided by $3$ if we compare the error on the $R_{\max}$ and the error on the residual risk after $n$ calls to the physical simulator. This is not surprising: the neural network makes a good job in ranking the contingencies but seems to have trouble identifying "how much" they are dangerous, especially for the most dangerous ones: the "extreme cases". The neural network seems to make a decent job in treating "average" cases, but to assess the risk of more dangerous contingencies, it is better to use a physical simulator. 

Even if our model is trained only on the single contingencies (eg less that $1\%$ of the grid states it is evaluated), the neural network is still able to accurately estimate the residual risk globally, provided that the impact of the most dangerous contingencies is quantified with a physical simulator.

\section{Conclusion}
In this paper, we proposed a novel approach to evaluate the dangerousness of a grid state with respect to some random events (in our case the unplanned disconnection of power lines). Results are evaluated on a standard benchmark. Our methodology can be summarized as follows: 

(1) Train a neural network to mimic a load flow simulator, on the data available.

(2) Use it (on new test data) to evaluate how close each line it to its thermal limits. We showed in this paper that even if the test data is drawn from a different distribution than the training data, this estimations works. Then rank contingencies in decreasing order of severity. 

(3) Estimate the risk of a simulations directly using machine learning, which allow great speed up in computational time, and thus to go beyond what is feasible today. 

(4) If a physical simulator is available, with \emph{almost no more computational cost} than what is done today, a better estimation is achievable by using the physical simulator on the worst contingencies, and relying on machine learning to estimate solely the least dangerous ones. In that case, even when facing unseen contingencies during training, the estimated residual risk is really close to the true one. Today the estimation of the risk over a lot of different events is difficult, as it requires too many computations if we rely purely on physical simulators.


Future work include detecting the amount of dangerous single contingencies, or adapting this framework in a wider area, where multiple grid states are evaluated at the same time. This could lead to rank contingencies from different grid states and could be used when studying forecasted grid states. 

\bibliographystyle{IEEEtran}

\bibliography{references}%

\end{document}